\documentclass[conference]{IEEEtran}
\IEEEoverridecommandlockouts
\usepackage{cite}
\usepackage{amsmath,amssymb,amsfonts}
\usepackage{algorithm}
\usepackage{algorithmic}
\usepackage{comment}
\usepackage{graphicx}
\usepackage{textcomp}
\usepackage{booktabs}
\usepackage{xcolor}
\usepackage{float} 
\usepackage{stfloats}
\usepackage{multirow}  
\usepackage{array}     

\usepackage{titlesec}
\usepackage[hang,flushmargin]{footmisc}

\titleformat{\paragraph}[runin]
{\normalfont\normalsize\itshape}{}{0pt}{}

\def\BibTeX{{\rm B\kern-.05em{\sc i\kern-.025em b}\kern-.08em
    T\kern-.1667em\lower.7ex\hbox{E}\kern-.125emX}}

\renewcommand{\thesubsubsection}{\Alph{subsection}.\arabic{subsubsection}}

\titleformat{\subsubsection}[runin]
{\normalfont\normalsize\bfseries}
{\thesubsubsection.}
{1em}
{}

\titlespacing{\subsubsection}{0pt}{10pt}{5pt}

\begin{document}

\title{GANGR: GAN-Assisted Scalable and Efficient Global Routing Parallelization
\thanks{This work was supported in part by the National Science Foundation under
Grant No. 2151854, titled End-to-End Global Routing with Reinforcement Learning in VLSI Systems.
}
}

\author{\IEEEauthorblockN{Hadi Khodaei Jooshin
}
\IEEEauthorblockA{\textit{Department of Electrical and Computer Engineering} 
\\
\textit{University of Illinois Chicago
}\\
Chicago, USA 
\\
hjoos@uic.edu
}
\and
\IEEEauthorblockN{Inna Partin-Vaisband
}
\IEEEauthorblockA{\textit{Department of Electrical and Computer Engineering
} \\
\textit{University of Illinois Chicago}
\\
Chicago, USA 
\\
vaisband@uic.edu
}
}

\maketitle

\begin{abstract}
Global routing is a critical stage in electronic design automation (EDA) that enables early estimation and optimization of the routability of modern integrated circuits with respect to congestion, power dissipation, and design complexity. Batching is a primary concern in top-performing global routers, grouping nets into manageable sets to enable parallel processing and efficient resource usage. This process improves memory usage, scalable parallelization on modern hardware, and routing congestion by controlling net interactions within each batch. However, conventional batching methods typically depend on heuristics that are computationally expensive and can lead to suboptimal results (oversized batches with conflicting nets, excessive batch counts degrading parallelization, and longer batch generation times), ultimately limiting scalability and efficiency. To address these limitations, a novel batching algorithm enhanced with Wasserstein generative adversarial networks (WGANs) is introduced in this paper, enabling more effective parallelization by generating fewer higher-quality batches in less time. The proposed algorithm is tested on the latest ISPD’24 contest benchmarks, demonstrating up to 40\% runtime reduction with only 0.002\% degradation in routing quality as compared to state-of-the-art router.
\end{abstract}


\section{Introduction}

Global routing algorithms target runtime efficiency, scalability (i.e., the ability to handle increasingly larger and more complex designs), and design quality (i.e., routability, congestion, power dissipation and wirelength). Existing approaches typically employ graph-based heuristics~\cite{liang2024ISPDGPUContest,liu2022fastgr}, iterative maze routing~\cite{liu2022fastgr,lin2022superfast}, and hierarchical decomposition~\cite{lin2024instantgr1,xiao2025instantgr2}, yielding improved yet limited routing performance. Achieving scalable routing requires more than improved algorithms---effective parallelization with hardware accelerators (GPUs, TPUs, etc.) is essential.

To parallelize global routing execution, traditional batching algorithms group nets that do not share routing resources into parallelizable batches. However, forming such batches is computationally expensive. For example, in InstantGR framework~\cite{lin2024instantgr1}, this process consumes up to 17.2\% of the overall routing runtime. Efficient batching is therefore a primary challenge in large, congested integrated circuits (ICs). Several studies have exploited GPU acceleration for maze routing, Steiner tree construction, and related subproblems \cite{lin2022GAMER,lin2022superfast,liu2023edge,liu2020CUGRDeatailRouting}, however, there hasn't been a GPU exploitation approach for batching nets using deep learning models.
For example, maze routing and net batching is accelerated with a GPU-based algorithm, GAMER~\cite{lin2022GAMER}, achieving a $2.7\times$ speedup compared to traditional CUGR~\cite{liu2020CUGRDeatailRouting}. GPU-accelerated FastGR router achieves a $2.489\times$ speedup compared with CUGR~\cite{liu2022fastgr}. A $13\times$ speedup over multithreaded CUGR is reported in~\cite{lin2022superfast} as a result of parallelization techniques. A scalable data partitioning and memory management schemes in the most recent InstantGR~\cite{xiao2025instantgr2} facilitate mapping of global routing tasks onto thousands of GPU threads, while balancing computational load across nets to avoid thread divergence. These improvements increase throughput compared to the CPU-based version, particularly on large three-dimensional (3D) benchmarks, positioning InstantGR as the top performing global router. A primary limitation with the existing GPU-based routers is their heavy reliance on time-consuming and inefficient heuristics used to achieve net-level parallelism. 
Thus, more efficient batching algorithms are required to reduce batch count, improve parallelism, and leverage GPU-like architectures more effectively.

In this paper, a batching framework for \textbf{fast generation} of \textbf{fewer and larger routing batches} is proposed. The key idea is to enhance parallelism by grouping a larger set of non-overlapping nets into fewer batches more effectively. The primary contributions are as follows:
\begin{itemize}
    \item Enhanced parallelism: a novel batching algorithm partitions nets into fewer larger batches, reducing inter-batch conflicts.

    \item Faster batching: WGAN~\cite{gulrajani2017WGANGP} is used to learn complex net-interference patterns, enabling faster and more accurate batch formation compared to existing heuristics. To the best of our knowledge, this is the first application of learning-based methods to routing batching.

    \item Runtime improvements: achieves faster global routing (due to WGAN-based accelerated batch generation and increased parallelism with fewer batches) with similar wirelength as compared with InstantGR. 
\end{itemize}

The rest of the paper is organized as follows.
The proposed framework is described in Section~\ref{sec:proposed}. The experimental results are presented in Section~\ref{sec:results}. The paper is concluded in Section~\ref{sec:conclusion}.

\section{Proposed Framework}\label{sec:proposed}
The proposed approach advances the parallelism principles introduced by InstantGR~\cite{lin2024instantgr1} in two key ways. First, the rule-based batching of nets is replaced with a WGAN-based mechanism trained to group non-overlapping nets more effectively, producing fewer and larger highly parallelizable batches. Second, CPU-only multithreaded batch processing is replaced with GPU acceleration, significantly increasing throughput. The WGAN-based batching algorithm is integrated into the first version of the InstantGR~\cite{lin2024instantgr1}, yielding up to 40\% faster runtime without additional overheads. 

\subsection{Overview of Net Overlap Analysis}
Net batching is a traditional approach for reducing global routing runtime~\cite{lin2022GAMER,liu2022fastgr,lin2022superfast,lin2024instantgr1,xiao2025instantgr2,chen2019DetailedRouting,chen2019DrCUDeatailRouting,chen2019DrCU2DeatailRouting,liu2020CUGRDeatailRouting}. The main idea is to parallelize the routing of nets that neither overlap nor share routing resources.
Owing to their simplicity, bounding boxes (i.e., the smallest rectangles enclosing the routing graphs of individual nets) are often used to detect potential net overlaps \cite{lin2022GAMER,liu2022fastgr,lin2022superfast,liu2020CUGRDeatailRouting}. This approach, however, has two main limitations: 1) it achieves a lower degree of parallelism, and 2) analyzing bounding box overlaps typically requires complex data structures (e.g., R-trees) that are difficult to optimize. 

To enhance parallelization, InstantGR routes nets simultaneously if their vertical segments do not overlap and their horizontal segments do not overlap, since these segments use different routing resources. However, it treats nets as overlapping whenever their segments are vertically or horizontally aligned, even if the segments are placed on different metal layers, thereby ignoring layer distinctions during parallelization.
%
To address this issue, metal layers are considered in this paper independently when analyzing overlaps, allowing nets to be routed in parallel, provided that no overlaps exist within the same metal layer.
To identify overlaps, the algorithm compares horizontal (vertical) segments only with horizontal (vertical) segments of other nets on the same metal layer. 

\begin{figure}[t!]
    \centering
    \includegraphics[width=\columnwidth]{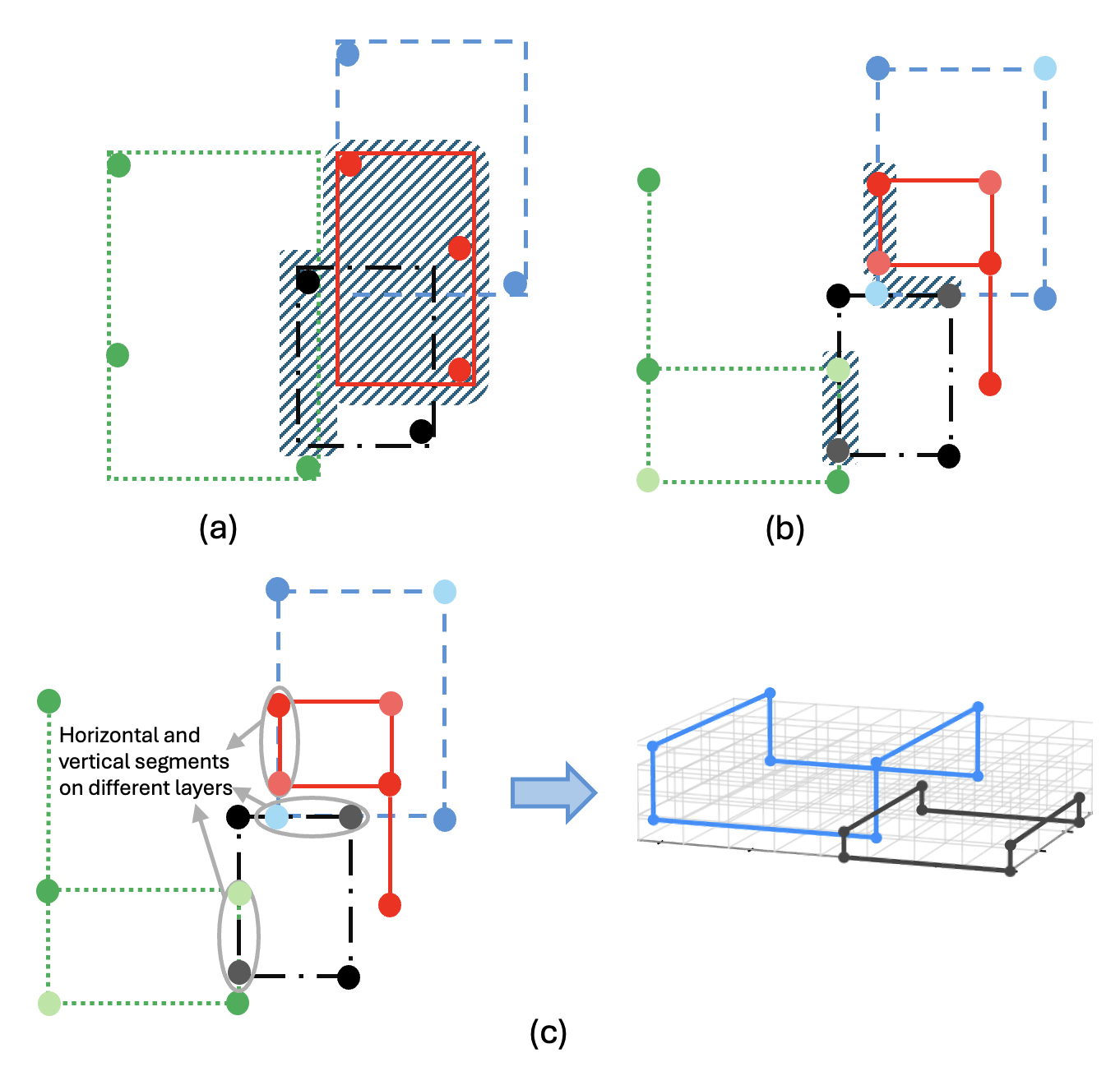}
    \caption{Net overlap analysis with conflicts (excluded from parallel routing) shown as hatched regions, (a) the bounding box method flags large overlap, limiting parallelism (used by many algorithms, e.g., CUGR2 \cite{liu2020CUGRDeatailRouting}, (b) the layer-agnostic analysis in InstantGR\cite{lin2024instantgr1}, which conservatively marks vertical and horizontal conflicts across all layers, underutilizing available parallelism, and (c) the proposed layer-aware method considers overlaps only within the same metal layer, avoiding false conflicts between nets on different layers, as illustrated by the black and blue nets in the 3D view.}
    \label{fig:overlapping_checking}
\end{figure}

These three batching approaches (i.e., bounding box-based, metal layer agnostic, and the proposed 3D-aware) are illustrated with four nets in Figure~\ref{fig:overlapping_checking}. In this case, the four nets cannot be parallelized with the bounding box and InstantGR batching approaches because of the overlaps in the bounding boxes (see Figure~\ref{fig:overlapping_checking}(a)) or the individual segments, i.e., horizontal for the black/blue nets and vertical for the green/black and red/blue nets, (see Figure~\ref{fig:overlapping_checking}(b)). Alternatively, if the vertical and horizontal segments of the nets that are considered overlapped by InstantGR are placed in different layers, routing of these nets can be parallelized (see Figure~\ref{fig:overlapping_checking}(c)). 

As previously noted, net overlap analysis is a primary concern in batching, since it is used both to construct batches and to detect conflicts that require reallocating overlapping nets. Consequently, overlap analysis directly impacts batching quality while also being computationally expensive. 
The proposed algorithm efficiently detects 3D spatial overlaps among nets within each batch, leading to improved batch validity and reduced conflicts.
Dynamic scheduling is used to distribute batch processing across multiple threads, where each thread maintains a local 3D visibility array~\footnote{a Boolean data structure indexed over the routing grid dimensions, where each entry indicates whether a grid cell is free (visible) or occupied, enabling constant-time conflict and overlap detection.} spanning the routing grid dimensions $(x_{G} \cdot y_{G} \cdot z_{G})$. For each net-batch pair, a layer-aware conflict detection is performed. 
If no conflicts are detected, the positions of all horizontal and vertical segments along with the pins are recorded in the visibility map to ensure proper isolation of routing elements across different metal layers in 3D space. 

\subsection{Wasserstein GAN-Based Layer-Aware Batching}
To the best of our knowledge, the proposed WGAN batching algorithm is the first algorithm that leverages GPUs using a deep learning-based approach. It operates in three stages.
\begin{enumerate}
    \item \textbf{WGAN-based initial batching} (see Algorithm~\ref{alg:ml_clustering}): groups $\sim$90\% of nets correctly (without overlaps). 
    \item \textbf{Adaptive evaluation of WGAN batches} (see Algorithm~\ref{alg:conflict_detection}): identifies conflicts within the individual batches based on layer-aware overlap criteria. 
    \item \textbf{Greedy layer-aware nets reallocation} (see Algorithm~\ref{alg:non_overlap_enforcement}): reallocates the remaining $\sim$10\% overlapping nets into suitable WGAN-generated batches, creating a new batch only when no fit exists.
\end{enumerate}

\subsubsection{Strategies: }\label{sec:strategies} The following strategies are utilized throughout the framework to optimize memory usage and facilitate synchronization.  

\paragraph{Data structures:} \hspace{1pt} 
By default, batches are stored in a dense 3D array, offering $\mathcal{O}(1)$ access time at the expense of increased memory usage, which is suitable for smaller ICs. When the total memory requirement $(\text{number of batches} \cdot x_{G} \cdot y_{G} \cdot l_{G})$ exceeds a predefined threshold, the algorithm switches to a sparse hash-based representation. Although slightly slower, this approach is far more memory-efficient and prevents overflow in large-scale designs while preserving overall efficiency.
Each hash set is pre-allocated for 1,000 elements that represent typical net occupancy patterns observed in industrial routing scenarios.

\paragraph{Memory management:} \hspace{1pt} Efficient data movement (i.e., C++ move semantics) and bulk operations are utilized to reduce memory allocation overhead.

\paragraph{Thread Synchronization:} \hspace{1pt} To maximize computational efficiency and avoid race conditions, parallel processing with carefully designed thread-local storage is employed for analyzing the individual batches for conflicts. Thread-local results are aggregated in shared data structures to avoid conflicting writes to shared memory.

\begin{algorithm}[b!]
\caption{WGAN-Based Initial Batching}
\label{alg:ml_clustering}
\begin{algorithmic}[1]
\small
\REQUIRE $nets$ (net IDs), $model$ (trained WGAN cached in GPU)
\ENSURE $batches$ (batched nets for parallel processing)

\STATE \textbf{Memory Management:} Compute adaptive chunk size: $\texttt{MAX\_CHUNK\_SIZE} \leftarrow \min(10^5, \max(10^4, |nets|/10))$
\FOR{each chunk of nets}
    \STATE \textbf{Feature Extraction:}
    \STATE \quad Extract normalized coordinates from net pin locations
    \STATE \quad Create 16-dimensional feature vector per net
    \STATE \quad Pad feature vector as needed

\STATE \textbf{Tensor Creation:} Convert feature vectors to PyTorch tensor and transfer to GPU memory
\STATE \textbf{ML Inference:} Net-batch distribution probability $P\!\leftarrow\!model$
\STATE \textbf{Batch Assignment:} For each net $n$, $n$'s batch index is $i~=~\arg\max(P)$, $batches[i] \leftarrow batches[i] + n$
\STATE \textbf{Memory Cleanup:} Transfer results to CPU; clear GPU tensor 
\ENDFOR
\STATE \textbf{Batch Aggregation:} Collect batch IDs from all chunks into unified result vector
\STATE \textbf{Batch Organization:} Group nets by predicted batch IDs with memory pre-allocated based on expected batch sizes to reduce dynamic allocation overhead
\STATE \textbf{Return:} $batches$ for subsequent parallel routing processing

\end{algorithmic}
\end{algorithm}

\begin{algorithm}[b!]
\caption{Adaptive Evaluation of WGAN Batches}
\label{alg:conflict_detection}
\begin{algorithmic}[1]
\small

\REQUIRE{$batches$ (batched nets), $grid\_dimensions$ $(x_{G}, y_{G}, l_{G})$}
\ENSURE{$accepted\_batches$ (conflict-free batches), $nets2reroute$ (conflicting nets)}

\FOR{each batch $i$ in $batches$ \textbf{in parallel}}
    \STATE Initialize $conflicts$ (conflict detection structure in dense or sparse representation; see Data structure strategy, Section~\ref{sec:strategies})
    \FOR{each pin $(x_p,y_p)$ of each net $n$ in batch $i$}
       
        \IF{\texttt{CONFLICT\_DETECTED}(($x_p,y_p$))}
            \STATE $nets2reroute \leftarrow nets2reroute + n$
        \ELSE
    \STATE $accepted\_batches \leftarrow accepted\_batches + n$
    \STATE Mark $conflicts[n.pins]$ as occupied
        \ENDIF
    \ENDFOR
\ENDFOR

\STATE Synchronize shared data structures across threads (see Thread synchronization strategy, Section~\ref{sec:strategies})


\STATE \textbf{Return:} $accepted\_batches$, $nets2reroute$

\vspace{5pt}

\STATE{\textbf{Function} \texttt{CONFLICT\_DETECTED}(pin ($x_p,y_p$))}
    \STATE \hspace{10pt} Extract layer $\ell\ \leftarrow \lfloor (x_p,y_p) / (x_G \cdot y_G) \rfloor$
    \STATE \hspace{10pt} $\text{pos}_\text{3D} \leftarrow \ell \cdot x_{G} \cdot y_{G} + x_p \cdot y_{G} + y_p$
    \STATE \hspace{10pt} Check conflicts in layer $\ell$; return Boolean yes/no

\end{algorithmic}
\end{algorithm}

\begin{algorithm}[b!]
\caption{Greedy Layer-Aware Nets Reallocation}
\label{alg:non_overlap_enforcement}
\begin{algorithmic}[1]
\small
\REQUIRE{$nets2reroute$, $\texttt{MAX\_BATCH\_SIZE}$, $batches$ (batched nets), $grid\_dimensions$ $(x_{G}, y_{G}, l_{G})$}
\ENSURE{$accepted\_batches$ (conflict-free batches with all nets assigned}


\FOR{each net $n$ in $nets2route$}
    \FOR{each batch $b$ in $batches$}
        \IF{$|b| + |batch\_assignments[b]| < \texttt{MAX\_BATCH\_SIZE}$ \AND !\texttt{CONFLICT\_DETECTED}$(n, b)$}
            \STATE $assigned\_batch \leftarrow b$
        \ENDIF
    \ENDFOR
    \IF{$not\_assigned\_batch$}
        \STATE $unassigned\_nets \leftarrow unassigned\_nets +n$
    \ELSE
        \STATE $batch\_assignments[assigned\_batch] \leftarrow batch\_assignments[assigned\_batch] + n$
    \ENDIF
\ENDFOR

\FOR{each batch $i$ in $batch\_assignments$ \textbf{in parallel}}
    \STATE Initialize $conflicts$ (conflict detection structure in dense or sparse representation; see Data structure strategy, Section~\ref{sec:strategies})
    \FOR{each pin $(x_p,y_p)$ of each net $n$ in batch $i$}
        \IF{\texttt{CONFLICT\_DETECTED}(($x_p,y_p$))}  
            \STATE $nets2reroute \leftarrow nets2reroute + n$
            \ELSE
            \STATE $accepted\_batches \leftarrow accepted\_batches + n$        
            \STATE Mark $conflicts[n.pins]$ as occupied
            \STATE Mark $conflicts[n.RSMT\_segmemts]$ as occupied 
        \ENDIF
    \ENDFOR
\ENDFOR

\STATE Synchronize shared data structures across threads (see Thread synchronization strategy, Section~\ref{sec:strategies})

\WHILE{$unassigned\_nets \neq \emptyset$}
    \STATE Exhaustively batch remaining nets into new batches until no more assignments possible
\ENDWHILE
\STATE Consolidate small batches ($\leq 5$ nets for datasets $< 10M$ nets, $\leq 10$ nets for datasets $> 10M$ nets) and report statistics
\end{algorithmic}
\end{algorithm}

\subsubsection{WGAN-Based Initial Batching 
(Algorithm~\ref{alg:ml_clustering}):}\label{sec:algo1}

The WGAN model is designed to efficiently batch nets in large-scale routing tasks. 
To optimize memory usage and computational efficiency with multithreading, nets are processed in chunks, with chunk size adaptively determined based on the total number of nets. For larger ICs, smaller chunks are used to prevent GPU out-of-memory issues and maintain scalability. This strategy enables processing of millions of nets within hardware limits. For each chunk, feature extraction (pin coordinates) and inference are performed sequentially. Extracted features are normalized by the grid dimensions to ensure consistent input scaling. Nets with fewer features than required are padded or repeated. All features are aggregated in a chunk into a feature matrix, converted into a tensor, and transferred to the GPU. The WGAN is used to predict batch assignments for all nets in the chunk, after which results are moved back to the CPU to free GPU memory for subsequent chunks. To capture inter-chunk dependencies, the predicted assignments for each net are stored in a global assignment list. 

Following this procedure, batches are generated with net indices assigned to each batch. The method enables scalable, memory-efficient batching by combining neural network inference, multithreading, and adaptive chunking.

\subsubsection{Adaptive Evaluation of WGAN Batches (Algorithm~\ref{alg:conflict_detection}):}\label{sec:algo2}
The WGAN-generated batches are evaluated for 3D-aware routing overlaps, with memory management strategies that adapt dynamically to problem size, ensuring scalability.
%
%
To enable efficient spatial queries across multi-layer routing paths, 
3D pin coordinates are mapped into unified spatial indices, based on the pin layer and its 2D position within the layer (see lines 15-16). 

Dense array access or sparse hash lookup are prototyped based on the preferred data structure strategy (see Section~\ref{sec:strategies}). 
The dense representation detects conflicts through direct array indexing with bounds checking, offering constant-time access but at the cost of high memory usage. In contrast, the sparse representation relies on hash set membership queries, which reduce memory overhead but introduce hashing latency. This trade-off favors dense methods for performance in heavily populated grids, while sparse methods scale better in sparse routing scenarios.

Per-thread data structures are maintained independently and the overall results are aggregated across threads in shared data structures using thread synchronization mechanisms.
To address variability in batch sizes, dynamic scheduling of parallel loops is employed to ensure load balancing across threads.



The adaptive approach ensures that the conflict detection overhead remains manageable even for large-scale industrial designs while maintaining the accuracy necessary for successful parallel routing execution, as demonstrated in the Results section.

\subsubsection{Greedy Layer-Aware Nets Reallocation (see Algorithm~\ref{alg:non_overlap_enforcement}):}\label{sec:algo3}
To resolve the spatial conflicts arising from the WGAN-based batching, conflicting nets are (i) identified and reassigned to conflict-free batches while preserving parallelization. The approach combines adaptive conflict resolution with efficient batch management to maximize routing throughput.

As a first step, a systematic assignment of conflicting nets into existing batches is attempted. 
Then, parallel processing with thread-local 3D arrays is employed to maintain isolation between concurrent evaluations.
%
All pins, along with the horizontal and vertical segments of the rectilinear Steiner minimal tree (RSMT) are recorded in the local arrays to ensure complete coverage of potential routing paths.

For nets that cannot be assigned without conflict within existing batches, an iterative greedy procedure is applied to generate additional conflict-free batches. The process is repeated until all nets are either assigned or identified as unassignable due to inherent constraints. Results from parallel processing phases are combined through thread synchronization mechanisms, ensuring data consistency and sustaining computational efficiency.


In the final phase, small batches are merged to reduce overhead: batches with at most five nets are combined for datasets containing up to 10M nets, and batches with at most ten nets are combined for larger datasets.


\subsection{ML Training with WGAN-Based Gradient Penalty}
An application-specific Wasserstein GAN with gradient penalty (WGAN-GP) architecture is designed for training the WGAN batching model. Training involves (i) data preprocessing, and (ii) adversarial learning under domain-specific constraints.

\subsubsection{Training Dataset Construction:}
The training dataset is generated by integrating the proposed layer-aware overlap batching criteria into the InstantGR algorithm~\cite{lin2024instantgr1}, and routing 177k nets from the NVLDA benchmark~\cite{liang2024ISPDGPUContest} with the adjusted InstantGR algorithm. The resulting output comprises pin coordinates, horizontal and vertical routing segments, and the batch number of each net. 

\subsubsection{Data Preprocessing and Feature Engineering:}
The training process begins with parsing routing batch files containing nets' pin coordinates, horizontal and vertical routing segments, half-perimeter wirelengths (HPWL), and batch assignments. To ensure statistical significance and maintain the number of batches consistent with the WGAN clustering output, only batches with more than 160 nets in the NVDLA benchmark are retained. 
The number of batches generated by the WGAN model (i.e., the WGAN batch number) is a key parameter influencing both batching quality and inference runtime. In this work, a fixed WGAN batch number is used across all evaluated benchmarks. The chosen value is experimentally determined to balance quality and runtime, as detailed in Section III.
A feature graph is then constructed, where each net is represented by a feature vector that incorporates pin coordinates. To reduce the WGAN input size, preprocessing time, and inference latency, only pin coordinates are used to construct the feature graph. Although this feature reduction lowers accuracy, it provides an acceptable tradeoff between runtime and accuracy. To improve the accuracy of the model at runtime cost, additional spatial characteristics can be included in the feature vector, such as bounding box dimensions, net centers, and vertical and horizontal segment perimeters. Additional preprocessing can also be added, including (i) feature detection through intersection-over-union (IoU) calculations, and (ii) segment overlap analysis to identify nets that cannot be assigned to the same batch due to resource conflicts. The size of the feature vector for each net is restricted to eight pins (i.e., 16 coordinates): nets with fewer than eight pins (i.e., about $92\%$ in the NVDLA benchmark) are fully represented, while for nets with more than eight pins, only the first eight pins are included in the feature graph.

\subsubsection{Training Loss Function:}
The proposed loss function balances three objectives.
The segment overlap loss penalizes nets within the same batch whose routing segments overlap, encouraging spatial separation. The center penalty loss discourages grouping nets with overlapping segments by applying a penalty proportional to the squared Euclidean distance between their pin centers, promoting compact batch formation. The pin overlap loss penalizes nets that share pins but assigned to the same batch, reducing pin congestion. Each term is weighted and normalized by the number of nets, yielding a differentiable loss that jointly enforces spatial separation, batch compactness, and balanced pin distribution for large-scale routing tasks. The proposed loss function is: 

\vspace{-10pt}
{\small
\begin{align}
\mathcal{L}_{\text{final}} & =  
\frac{w_{\text{seg}}}{N} \sum_{(i, j) \in \mathcal{N}_{seg}} \left( \sum_{b=1}^{B} \left(p_i^b p_j^b\right)^2 \right) \nonumber \\
&+ \frac{w_{\text{ctr}}}{N} \sum_{(i, j) \in \mathcal{N}_{seg}} \left( \sum_{b=1}^{B} \left(p_i^b p_j^b\right) \|c_i - c_j\|^2 \right) \nonumber \\
&+ \frac{w_{\text{pin}}}{N} \sum_{p \in \mathcal{P}} \sum_{(i,j)\in\mathcal{N}_p} \left(\sum_{b=1}^{B} \left(p_i^b p_j^b\right)^2 \right).
\label{eq:clustering_loss}
\end{align}
}
\begin{itemize}
    \scriptsize
    \item $N$: Total number of nets in the dataset.
    \item $w_{\text{seg}}$, $w_{\text{ctr}}$, $w_{\text{pin}}$: Weighting coefficients for segment overlap, center penalty, and pin overlap loss terms, respectively.
    \item 
    $\mathcal{N}_{seg}$: Set of pairs of nets with overlapping routing segments, as defined by the conflict graph.
    \item 
    $\mathcal{P}$: Set of all unique pin locations in the design.
    \item 
    $\mathcal{N}_p$: Set of pairs of nets that share the same pin $p \in \mathcal{P}$ location.
    \item $B$: Total number of batches 
    \item $p_i^b$: Assignment probability of net $i$ to batch $b$, as predicted by the WGAN model.
    \item $c_i$: Center coordinate (mean pin location) of net $i$.
\end{itemize}

\begin{table}[t!]
\centering
\caption{Benchmark Details [ISDP2024]~\cite{liang2024ISPDGPUContest}.}
\vspace{-5pt}
\label{tab:benchmark_details}
\scriptsize
\setlength{\tabcolsep}{4pt}
\begin{tabular}{clrrr}
\toprule
\textbf{IC} & \textbf{Benchmark} & \textbf{\#Nets} & \textbf{\#Pins} & \textbf{Gcell Grid} \\
\midrule
0 & Ariane\_sample & 129K & 420K & $844 \times 1144$ \\
1 & MemPool-Tile\_sample & 136K & 500K & $475 \times 644$ \\
2 & NVDLA\_sample & 177K & 630K & $1240 \times 1682$ \\
3 & BlackParrot\_sample & 770K & 2.9M & $1532 \times 2077$ \\
4 & MemPool-Group\_sample & 3.3M & 10.9M & $1782 \times 2417$ \\
5 & MemPool-Cluster\_sample & 10.6M & 40.2M & $3511 \times 4764$ \\
6 & TeraPool-Cluster\_sample & 59.3M & 213M & $7891 \times 10708$ \\
7 & Ariane\_rank & 128K & 435K & $716 \times 971$ \\
8 & MemPool-Tile\_rank & 136K & 483K & $429 \times 581$ \\
9 & NVDLA\_rank & 174K & 610K & $908 \times 1682$ \\
10 & BlackParrot\_rank & 825K & 2.9M & $1532 \times 2077$ \\
11 & MemPool-Group\_rank & 3.2M & 10.9M & $1782 \times 2417$ \\
12 & MemPool-Cluster\_rank & 10.6M & 40.2M & $4113 \times 5580$ \\
13 & TeraPool-Cluster\_rank & 59.3M & 213M & $9245 \times 12544$ \\
\bottomrule
\end{tabular}
\end{table}

\subsubsection{Adversarial Training with Spatial Constraints:}
The WGAN framework consists of a generator and a critic. The generator is a 5-layer fully connected network with residual connections, LeakyReLU activations, and layer normalization, producing soft clustering probability distributions for input nets. The critic is an 8-layer network designed for spatial feature extraction, distinguishing generator outputs from reference clustering patterns. Training alternates between five critic updates and one generator update, following the WGAN-GP protocol. The critic loss maximizes the Wasserstein distance\footnote{a metric used to measure the dissimilarity between the distribution of real data and the distribution of generated data}~\cite{gulrajani2017WGANGP} between real and generated assignments with a gradient penalty ($\lambda_{gp} = 10.0$) to enforce the Lipschitz constraint\footnote{a mathematical condition applied to the critic function}. The generator optimizes a composite objective combining adversarial loss with domain-specific constraints: segment overlap (routing conflicts), pin overlap (resource contention), center distance (spatial locality), and cluster balance, with tuned loss weights. Both networks are trained with Adam ($\text{learning rate} = 0.0003$, $\beta_1 = 0.0$, $\beta_2 = 0.9$). Early stopping based on overlap reduction metrics prevents overfitting and ensures convergence to physically realizable batching solutions.

\begin{table}[t!]
\centering
\caption{Score and Runtime Performance of GANGR Compared With InstantGR~\cite{lin2024instantgr1}}
\vspace{-5pt}
\label{tab:GANGR_vs_IGR}
\scriptsize
\setlength{\tabcolsep}{4pt}
\begin{tabular}{c|cc|cc}
\hline
\multirow{2}{*}{\textbf{Bench.}} & \multicolumn{2}{c|}{\textbf{InstantGR}} & \multicolumn{2}{c}{\textbf{GANGR}} \\
& \textbf{Score} & \textbf{Time (s)} & \textbf{Score} & \textbf{Time (s)} \\
\hline
0 & 19715069 & 1.51 & 19714965 & 1.51 \\
1 & 15124499 & 1.67 & 15126701 & 1.28\\
2 & 47979984 & 2.65 & 48038696 & 1.74 \\
3 & 112468847 & 9.44 & 112463498 & 6.82 \\
4 & 397600677 & 27.83 & 398550511 & 15.24 \\
5 & 1623738805 & 90.30 & 1644464118 & 59.21 \\
7 & 22544301 & 1.82 & 22539569 & 1.58 \\
8 & 13772197 & 1.62 & 13780323 & 1.27 \\
9 & 43044750 & 2.64 & 43140388 & 2.14 \\
10 & 109840647 & 5.73 & 109855333 & 4.33 \\
11 & 382576253 & 23.42 & 383171693 & 15.06 \\
12 & 1780759982 & 97.65 & 1788086646 & 59.58 \\
\hline
\textbf{Avg. Ratio} & \textbf{0.998} & \textbf{1.395} & \textbf{1.000} & \textbf{1.000} \\
\hline
\end{tabular}
\end{table}

\section{Experimental Results}\label{sec:results}
The proposed algorithms are integrated into the InstantGR framework~\cite{lin2024instantgr1}. Experiments are performed on a 64-bit Linux workstation with an AMD Ryzen 9 7950X 16-core processor, 124 GB of memory, and a single NVIDIA GeForce RTX 4090 using 8 CPU threads. The ISPD’24 global routing contest~\cite{liang2024ISPDGPUContest} benchmarks (see Table~\ref{tab:benchmark_details}) and evaluator~\cite{liang2024ISPDGPUContest} are used to evaluate the existing \cite{lin2024instantgr1} and proposed frameworks. 
Performance is measured with the ISPD’24 contest metric---a weighted cost function combining total wirelength, via count, and overflow penalty~\cite{liang2024ISPDGPUContest}.

The score and runtime performance of GANGR compared with InstantGR~\cite{lin2024instantgr1} is presented in Table~\ref{tab:GANGR_vs_IGR}. For a fair comparison, InstantGR, the only publicly available version among the current SOTA frameworks, was re-evaluated on the previously described local workstation together with GANGR.
As compared with InstantGR, the proposed GANGR framework achieves $\sim$40\% improvement in runtime with only 0.002\% reduction in routing score. GANGR achieves speedup through both efficient batch generation and parallelization. On benchmark \#12, segment-based InstantGR produces 383 batches in 9.3 s, while GANGR generates only 68 batches (82\% fewer) in 4.4 s (52\% faster), enabling greater parallelization. 

A detailed comparison of GANGR with InstantGR is presented in Table~\ref{tab:detailed_comparison}. The average wirelength of GANGR matches that of InstantGR. The average via count is 0.001\% higher than InstantGR, while the overflow is 0.008\% higher. Given the runtime improvement achieved, these small degradations are considered negligible.

\begin{table*}[t]
\centering
\caption{Wirelength, Via Count, and Overflow Performance of GANGR Compared to InstantGR~\cite{lin2024instantgr1}.}
\vspace{-5pt}
\label{tab:detailed_comparison}
\scriptsize
\setlength{\tabcolsep}{4pt}
\begin{tabular}{c|cc|cc|cc}
\hline
\multirow{2}{*}{\textbf{Bench}} & \multicolumn{2}{c|}{\textbf{Wirelength}} & \multicolumn{2}{c|}{\textbf{Via Count}} & \multicolumn{2}{c}{\textbf{Overflow}} \\
& \textbf{InstantGR} & \textbf{GANGR} & \textbf{InstantGR} & \textbf{GANGR} & \textbf{InstantGR} & \textbf{GANGR} \\
\hline
0 & 9361015 & 9361764 & 2794988 & 2798052 & 7560742 & 7555149 \\
1 & 8350182 & 8352319 & 3259740 & 3262132 & 3516588 & 3512251 \\
2 & 21289207 & 21279195 & 4302208 & 4301116 & 22390810 & 22458385 \\
3 & 58157063 & 58158384 & 18932720 & 18935032 & 35385097 & 35370082 \\
4 & 260156132 & 260428582 & 71296860 & 71485016 & 66205020 & 66636913 \\
5 & 1091435102 & 1092394481 & 256292732 & 256701240 & 276218463 & 295368397 \\
7 & 11972097 & 11973312 & 2827892 & 2828236 & 7745811 & 7738022 \\
8 & 7553701 & 7555546 & 3212488 & 3217760 & 3008600 & 3007016 \\
9 & 21582358 & 21578627 & 4411648 & 4408724 & 17053141 & 17153037 \\
10 & 55660842 & 55665285 & 18993980 & 18999924 & 35189233 & 35190124 \\
11 & 248207884 & 248426271 & 70609280 & 70845020 & 63822725 & 63900402 \\
12 & 1189011781 & 1189873247 & 267495852 & 267769824 & 324389756 & 330443575 \\

\hline
\textbf{Avg. Ratio} & \textbf{1.000} & \textbf{1.000} & \textbf{0.999} & \textbf{1.000} & \textbf{0.992} & \textbf{1.000} \\
\hline
\end{tabular}
\end{table*}

\begin{table*}[t]
\centering
\caption{Score and Runtime Performance of GANGR as a Function of the WGAN batch number.}
\vspace{-5pt}
\scriptsize
\label{tab:GANGR_result_initial_cluster_numbers}
\setlength{\tabcolsep}{4pt}
\begin{tabular}{c|cc|cc|cc|cc|cc}
\hline
\multirow{2}{*}{Benchmark} & \multicolumn{2}{c|}{$n = 20$} & \multicolumn{2}{c|}{$n = 30$} & \multicolumn{2}{c|}{$n = 40$} & \multicolumn{2}{c|}{$n = 50$} & \multicolumn{2}{c}{$n = 60$} \\
& Score & Time (s) & Score & Time (s) & Score & Time (s) & Score & Time (s) & Score & Time (s) \\
\hline
0 & \textbf{19712211} & \textbf{1.35} & 19714965 & 1.51 & 19715790 & 1.69 & 19716959 & 1.85 & 19719862 & 2.03\\
1 & \textbf{15119754} & \textbf{1.16} & 15126701 & 1.28 & 15130345 & 1.41 & 15132497 & 1.52 & 15129001 & 1.62\\
2 & 48176669 & 2.08 & 48038696 & \textbf{1.74} & 48028344 & 1.89 & 48008122 & 2.02 & \textbf{48007290} & 2.17\\
3 & 112530612 & \textbf{6.35} & 112463498 & 6.82 & \textbf{112461801} & 7.12 & 112479244 & 7.42 & 112468346 & 7.83 \\
4 & 400002097 & 15.45 & 398550511 & 15.24 & 398189929 & \textbf{15.17} & 398241192 & 15.23 & \textbf{398141390} & 15.52 \\
5 & 1755161764 & 61.8 & 1644464118 & 59.21 & 1645831061 & \textbf{58.35} & 1631026125 & 58.73 & \textbf{1627691371} & 59.61 \\
7 & 22542373 & \textbf{1.44} & 22539569 & 1.58 & 22538789 & 1.77 & \textbf{22536818} & 1.92 & 22541876 & 2.1\\
8 & 13764719 & \textbf{1.18} & 13780323 & 1.27 & 13773858 & 1.41 & \textbf{13763651} & 1.48 & 13767970 & 1.58\\
9 & 43270730 & 1.84 & 43140388 & 2.14 & 43107865 & \textbf{1.77} & 43113495 & 1.89 & \textbf{43062996} & 1.99\\
10 & \textbf{109827103} & \textbf{4.2} & 109855333 & 4.33 & 109852896 & 4.53 & 109857836 & 4.73 & 109856100 & 4.91\\
11 & 383748478 & \textbf{14.88} & 383171693 & 15.06 & 383049653 & 15.01 & 383069104 & 15.22 & \textbf{383016832} & 15.38\\
12 & 1949165122 & 61.54 & 1788086646 & 59.58 & 1789137454 & \textbf{59.02} & 1786770109 & 59.66 & \textbf{1782743886} & 59.11\\
\hline
Avg. Ratio & \textbf{1.000} & \textbf{1.000} & \textbf{0.987} & \textbf{1.035} & \textbf{0.987} & \textbf{1.071} & \textbf{0.986} & \textbf{1.124} & \textbf{0.986} & \textbf{1.182}\\
\hline
\end{tabular}
\end{table*}
As noted earlier, in this work, a fixed WGAN batch number is used across all evaluated benchmarks.
Experimental results for different WGAN batch number are summarized in Table~\ref{tab:GANGR_result_initial_cluster_numbers}. Increasing the number of batches reduced the average score while increasing runtime. However, the strong postprocessing stage mitigates the effect of fewer batches on quality, resulting in only a minor score reduction. Thus, a smaller WGAN batch number is preferred. To maintain consistency across experiments, 30 batches ($n = 30$) is selected as it offers the best balance between quality and runtime. While assigning benchmark-specific values could further optimize results, this option is not considered here for fair comparison.

\section{Conclusion}\label{sec:conclusion}
In this paper, GANGR is introduced, a deep learning–assisted batching framework that integrates Wasserstein generative adversarial networks (WGANs) with a newly developed batching approach in the global routing pipeline. Unlike heuristic-based batching methods, the proposed framework learns complex net-interference patterns and partitions nets into fewer batches, thereby improving parallelism and scalability. With layer-aware overlap detection and adaptive memory-efficient validation strategies, preprocessing and batch generation time are reduced while routing quality is maintained across large-scale industrial benchmarks. Experimental results on the ISPD’24 benchmarks show that GANGR achieves runtime improvements of up to 40\% over state-of-the-art global routers, while maintaining competitive wirelength, via count, and congestion metrics.

\bibliographystyle{plain}
\bibliography{myreferences} 
\end{document}